\documentclass[letterpaper, 10 pt, conference]{ieeeconf}

\IEEEoverridecommandlockouts
\usepackage{graphicx}
\usepackage{amsmath}
\usepackage{amssymb}
\usepackage{booktabs}
\usepackage{float}
\usepackage{url}

\title{\LARGE \bf
From Legs to Wheels: Embodiment-Aware Human Motion Retargeting for Mobile-Base Humanoids
}

\author{Jiyeon Koo, Eunseom Pyo, Jeonghee Seo, Taehwa Kim, and Andrew Jaeyong Choi$^{*}$%
\thanks{The authors are with iRASC Lab., School of Computing, Gachon University, Seongnam 13120, Republic of Korea. Emails: \{halo1225, vydms123, sjh0621, tempteru, andrewjchoi\}@gachon.ac.kr.}%
\thanks{$^{*}$Corresponding author: Andrew Jaeyong Choi.}%
}

\begin{document}

\maketitle
\thispagestyle{empty}
\pagestyle{empty}

\begin{abstract}
Human video offers a scalable source of robot demonstrations, yet most human-to-humanoid retargeting methods assume a legged robot with human-like kinematics. This assumption does not hold for mobile-base humanoids equipped with a wheeled base, vertical lift, and two arms. Human walking must be expressed through base motion, while torso bending may require coordinated lift and arm motion. We address this mismatch with a task-conditioned framework that assigns reconstructed human motion to base, lift, and arm responsibilities before robot-specific realization. The allocator preserves the human-derived path, stabilizes heading, separates turn and translation when needed, retimes commands to satisfy base limits, and repairs lift and arm trajectories. A deployment adapter then converts the reference to 50 Hz commands using stationary-base detection, deadband and slew-rate filtering, time-consistent playback scaling, and separate linear and angular gains. We evaluate the resulting references with human-derived task-space comparisons, policy-free simulation replay, and a qualitative execution on a physical robot.
\end{abstract}

\section{INTRODUCTION}

Humanoid motion imitation has advanced through the combination of human motion capture, retargeting, and reinforcement learning (RL). Recent systems retarget whole-body motion to legged humanoids and train policies to follow the resulting references in simulation or on hardware~\cite{cheng2024exbody,fu2024humanplus,he2024omnih2o}. These pipelines generally assume that the target is also a biped. Human walking, torso orientation, and limb motion can therefore be mapped to robot legs, trunk, and arms with fairly direct correspondence.

We consider a different embodiment: \emph{mobile-base humanoids}. These robots retain a human-facing upper body and two arms, but replace the legs with a wheeled base and may use a vertical lift in place of a human-like spine. Simple joint correspondence is no longer sufficient. A person walking forward with still arms should induce base motion, whereas bending to reach a low object may require changes in base position, lift height, and arm configuration. Retargeting must therefore determine not only how to reproduce a motion, but also which robot subsystem should express each human cue.

\begin{figure}[!t]
    \centering
    \includegraphics[width=\columnwidth]{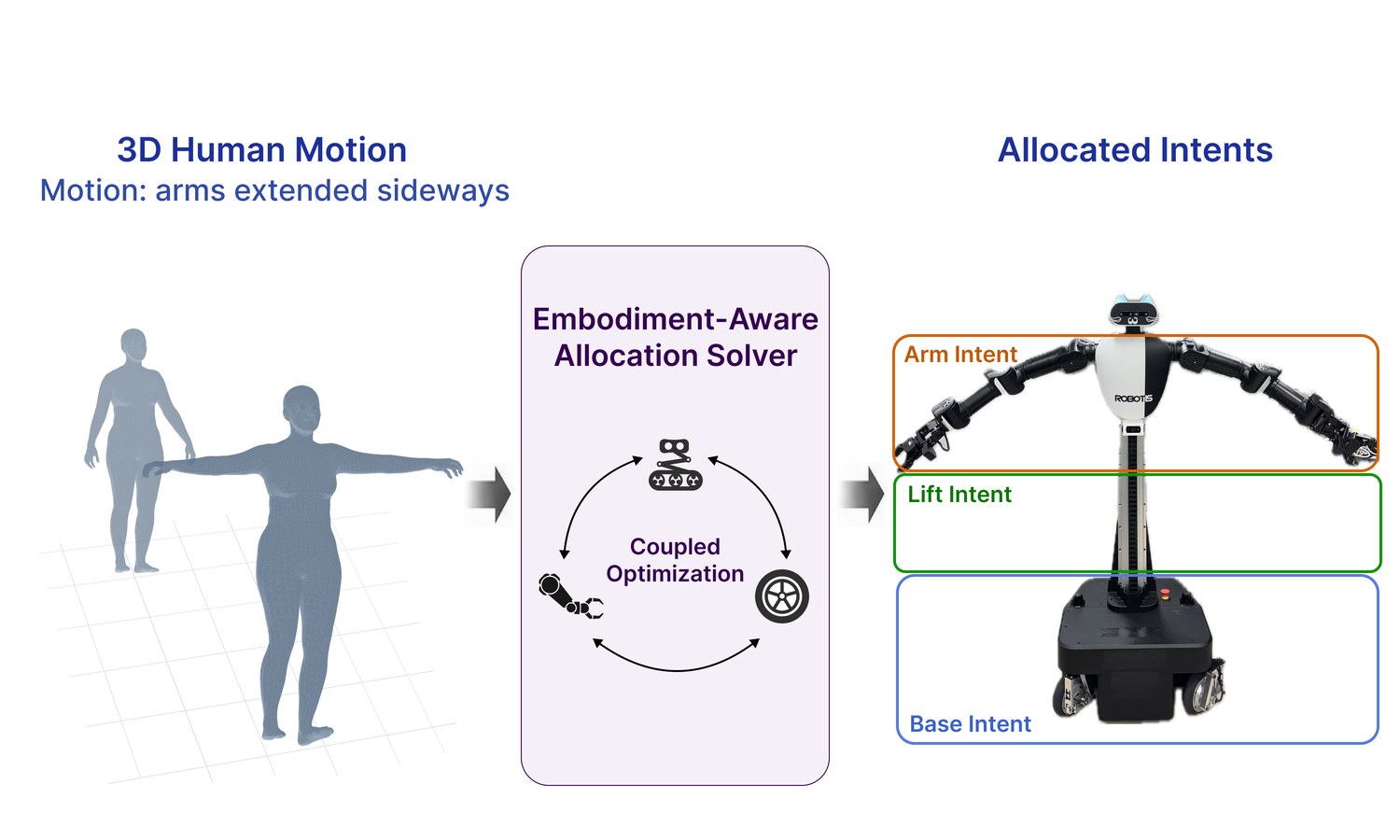}
    \caption{Embodiment-aware allocation of reconstructed human motion into arm, lift, and mobile-base intents. A single whole-body motion may require coordinated actuation across several robot subsystems.}
    \label{fig:overview}
\end{figure}

\begin{figure*}[!t]
    \centering
    \includegraphics[width=\textwidth]{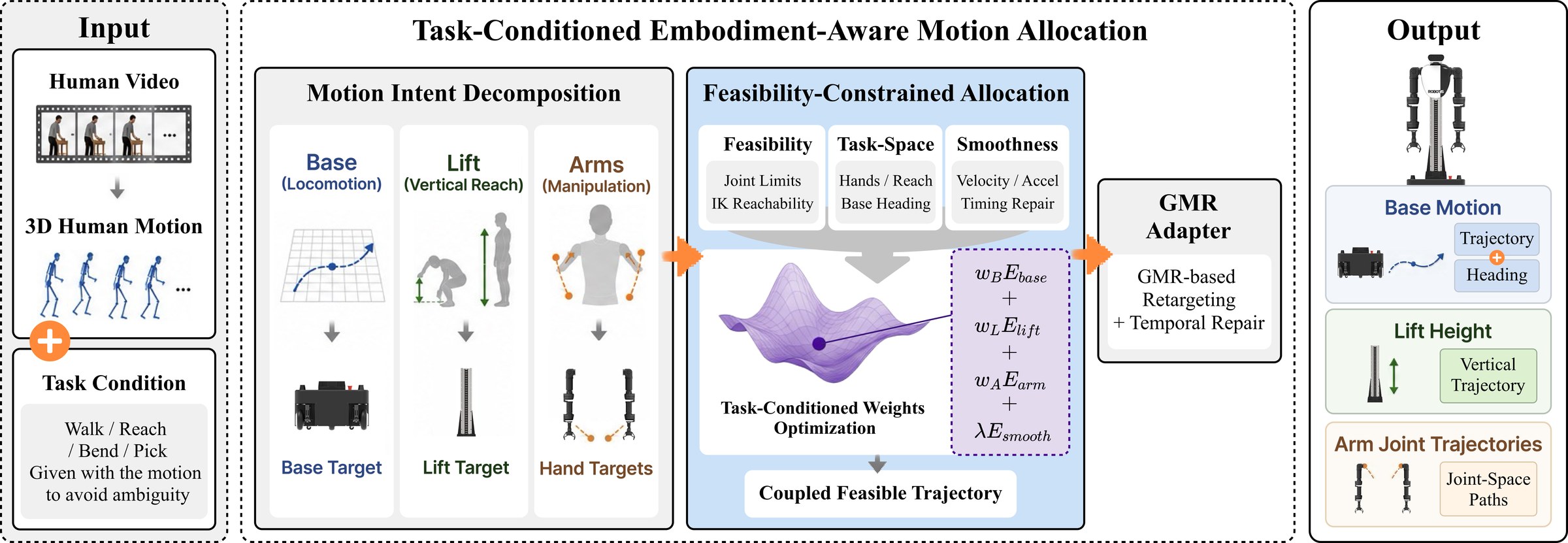}
    \caption{Task-conditioned embodiment-aware retargeting pipeline. Human video is reconstructed as 3D motion and decomposed into base, lift, and arm targets. These targets are optimized under feasibility, task-space, and smoothness objectives, then realized and repaired by a GMR-based adapter for simulation playback or command export.}
    \label{fig:architecture}
\end{figure*}

Fig.~\ref{fig:overview} illustrates the embodiment mismatch, and Fig.~\ref{fig:architecture} gives an overview of our approach. From a reconstructed motion sequence and a lightweight task condition, the framework extracts locomotion intent for the base, vertical reach and body-height intent for the lift, and manipulation intent for the arms. A robot-specific realization module converts these components into commands that preserve task-space behavior within kinematic and command limits.

The allocation formulation does not depend on a particular 3D human-motion reconstruction method. We use GMR for robot-specific upper-body realization and study retargeting quality, reference feasibility, simulation playback, and initial physical-robot execution. Learning-based tracking performance is outside the scope of this work.

The contributions are:
\begin{itemize}
    \item A task-conditioned motion allocator with explicit base, lift, and arm variables. Phase-gated human-motion cues replace one-to-one joint matching.
    \item A subsystem-specific temporal repair method that preserves base-path progress, stabilizes heading, decouples turn and translation, retimes motion to platform limits, and coordinates lift and arm trajectories.
    \item A 50 Hz deployment adapter for time-consistent command conversion, together with evaluation using human-derived task-space costs, component ablations, policy-free simulation replay, and qualitative physical-robot execution.
\end{itemize}

\section{RELATED WORK}

\subsection{Monocular Human Motion Reconstruction}

Human-video imitation typically begins by converting RGB observations into a body representation such as SMPL~\cite{loper2015smpl}. VIBE improves temporal consistency in pose and shape estimates~\cite{kocabas2020vibe}, while HuMoR uses a motion prior to handle noisy or missing observations~\cite{rempe2021humor}. More recent monocular methods recover world-grounded root motion in addition to local body pose. GLAMR estimates globally consistent motion under dynamic cameras and long occlusions~\cite{yuan2022glamr}, and SLAHMR jointly models human and camera trajectories~\cite{ye2023slahmr}. WHAM refines global motion using contact information~\cite{shin2024wham}. TRAM combines camera trajectory recovery with in-the-wild body motion estimation~\cite{wang2024tram}, whereas GVHMR uses gravity-view coordinates to reduce ambiguity and drift~\cite{shen2024gvhmr}. We use this line of work for RGB-to-3D reconstruction, then allocate the recovered motion to the actuation space of a mobile-base humanoid.

\subsection{Human-to-Humanoid Motion Retargeting}

Classical whole-body retargeting transfers human link poses to analogous robot links through geometric task objectives and inverse kinematics~\cite{darvish2019geometric}. GMR and related tracking pipelines show that reference quality directly affects downstream policy learning and tracking robustness~\cite{araujo2025retargeting}. Infeasible poses and inconsistent contacts should therefore be corrected before control. Existing methods, however, mainly target legged humanoids and represent human locomotion through robot leg motion. Our setting instead has to express that locomotion with a mobile base.

RL-based imitation has evolved from reference tracking in DeepMimic~\cite{peng2018deepmimic} to reusable motion priors and skill embeddings learned from large motion collections~\cite{peng2021amp,peng2022ase}. PHC~\cite{luo2023phc}, H2O~\cite{he2024h2o}, ExBody~\cite{cheng2024exbody}, HumanPlus~\cite{fu2024humanplus}, OmniH2O~\cite{he2024omnih2o}, and BeyondMimic~\cite{truong2025beyondmimic} demonstrate whole-body tracking of retargeted references. Recent Humanoids work also learns poses from internet video~\cite{mao2025universal} and generates editable motion through postural synergies~\cite{malhotra2025scripting}. Our focus is earlier in the pipeline: generating feasible references when the robot's lower body is not human-like.

\subsection{Mobile Manipulation from Human Demonstrations}

Mobile ALOHA learns bimanual mobile manipulation from whole-body teleoperation data~\cite{fu2024mobilealoha}. Related Humanoids studies cover compact mobile humanoid hardware~\cite{kang2025anubis}, non-holonomic whole-body impedance control~\cite{forouhar2025garmi}, and coupled versus decoupled embodiment in whole-body teleoperation~\cite{moyen2025embodiment}. Human demonstrations have also supported manipulation-speed adaptation~\cite{tsunoda2025velocity} and assistive motion generation~\cite{arai2025sitting}. WARP retargets offline demonstrations to whole-body mobile manipulation while accounting for embodiment differences~\cite{chen2026warp}. Our work asks a complementary question: how should reconstructed whole-body cues be divided among a mobile base, vertical lift, and two arms before GMR-based realization?

\subsection{Optimization-Based Retargeting and IK}

Retargeting is commonly formulated as constrained optimization over robot configurations. The objective minimizes task-space error subject to joint, velocity, collision, and smoothness constraints. Differential inverse kinematics tools such as Mink solve weighted velocity-level tasks in a constrained MuJoCo model~\cite{zakka2025mink}. Recent Humanoids work learns differentiable reachability constraints for motion generation~\cite{murooka2025reachability} and uses continuous contact optimization for whole-body manipulation~\cite{leve2025contact}. We add an allocation stage before GMR realization to decide whether each human cue should drive the base, lift, arms, or a combination of these subsystems.

\section{PROBLEM FORMULATION}

Let a human video be converted into a time-indexed motion representation
\begin{equation}
    H_{1:T} = \{h_t\}_{t=1}^{T},
\end{equation}
where $h_t$ contains the estimated pelvis/root pose, torso orientation, shoulder and wrist positions, available hand orientations, and optional object/contact cues. The source of $H_{1:T}$ is not fixed. It may be a monocular estimator, a motion-capture system, or an existing 3D motion file.

The sequence is paired with a task condition
\begin{equation}
    c = \{\text{task mode}, \text{allowed phases}, \text{object cues}\},
\end{equation}
where the task mode specifies locomotion, manipulation, carry, or a general mode. This condition does not perform semantic action recognition. It instead controls how the optimizer weights human cues. For example, arm swing during locomotion should not become a primary manipulation target, whereas hand motion should receive greater weight in a manipulation task.

The target robot is modeled as
\begin{equation}
    R = \{B, L, A_l, A_r\},
\end{equation}
where $B$ is the mobile base, $L$ is the vertical lift or body-height mechanism, and $A_l, A_r$ are the left and right arms.

The robot command at time $t$ is
\begin{equation}
    u_t = [v^x_t, v^y_t, \omega_t, \dot{\ell}_t, \dot{q}^{\,l}_t, \dot{q}^{\,r}_t],
\end{equation}
or the corresponding position targets, depending on the low-level controller. The goal is to find a command sequence $U_{1:T}$ and allocation sequence $Z_{1:T}$ that make the robot imitate the \emph{function} of the human motion:
\begin{align}
    (U_{1:T}^{*},Z_{1:T}^{*})
    &= \arg\min_{U_{1:T},Z_{1:T}} \mathcal{J}(U,Z;H,c,R), \notag \\
    \mathcal{J}
    &= E_{\text{task}}(U,Z;H,c) + E_{\text{alloc}}(Z;H,c,R) \notag \\
    &\quad + E_{\text{feas}}(U;R) + E_{\text{smooth}}(U,Z).
    \label{eq:objective}
\end{align}

Here $E_{\text{task}}$ penalizes deviations in hand trajectories, approach direction, reach height, and temporal phase. $E_{\text{alloc}}$ encourages a consistent assignment of human motion components to robot subsystems. $E_{\text{feas}}$ penalizes joint-limit, workspace, velocity, acceleration, and collision violations. $E_{\text{smooth}}$ regularizes base, lift, and arm commands over time.

The key difference from conventional retargeting is the allocation variable
\begin{equation}
    z_t = [z^B_t, z^L_t, z^{A_l}_t, z^{A_r}_t],
\end{equation}
which encodes the responsibility of each robot subsystem at time $t$. Human pelvis translation should mainly activate $z^B_t$, whereas vertical reach should activate both $z^L_t$ and the relevant arm component. Forward bending may be distributed across base approach, lift adjustment, and arm extension.

\section{METHOD}

Table~\ref{tab:task_index} lists the scenarios used throughout the paper, the motion cue emphasized in each one, and the intended subsystem allocation.

\begin{table*}[!t]
    \caption{TASK INDEX AND INTENDED EMBODIMENT-AWARE MOTION ALLOCATION.}
\label{tab:task_index}
\centering
\scriptsize
\setlength{\tabcolsep}{2pt}
\begin{tabular}{clp{0.25\textwidth}p{0.25\textwidth}p{0.20\textwidth}}
\toprule
ID & Task & Description & Primary Motion Cue & Allocation Target \\
\midrule
01 & Slow walk & Forward walking at a moderate pace & Smooth root translation & Base \\
02 & Run & Fast forward locomotion & High root velocity & Base (speed-adaptive) \\
03 & Side walk & Lateral walking without a dominant turn & Lateral root motion & Lateral base motion \\
04 & Sideways-turn walk & Lateral locomotion combined with reorientation & Root translation + yaw & Base translation + rotation \\
05 & Bend-and-reach I & Forward reach with moderate torso lowering & Torso pitch + low wrist & Lift + arms \\
06 & Bend-and-reach II & Deeper forward reach at a lower height & Deep reach + arm extension & Lift + arms + base assist \\
07 & Backward step with arms forward & Base retreats while both hands remain directed forward & Opposing root--wrist motion & Base + coordinated arms \\
08 & Wide-arm lateral motion & Broad lateral arm movement during body relocation & Wide bilateral reach & Base + arms \\
09 & Low-object pick & Approach and close around a low target & Low wrist + grasp approach & Lowered lift + arms \\
\bottomrule
\end{tabular}
\end{table*}

\begin{figure}[H]
    \centering
    \includegraphics[width=\linewidth]{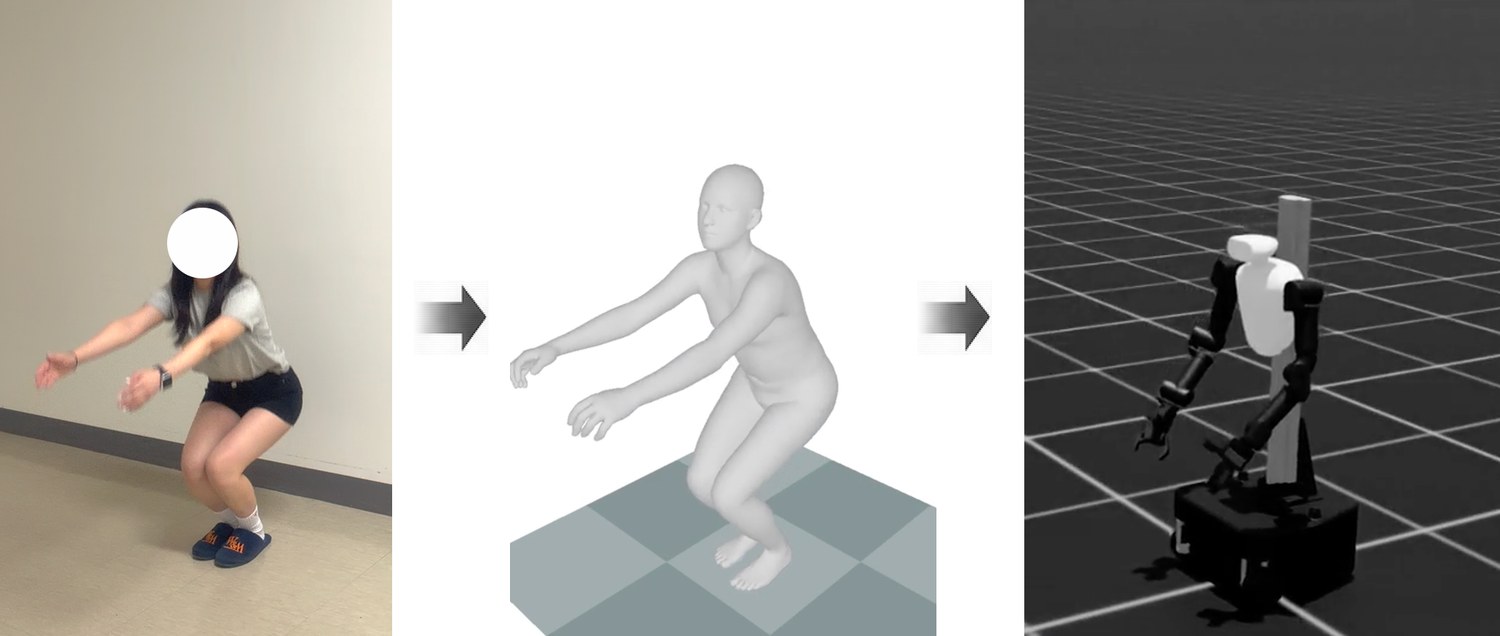}
    \caption{Example processing sequence for a bend-and-reach motion: RGB human demonstration, GVHMR 3D reconstruction, and the retargeted mobile-base humanoid reference in simulation.}
    \label{fig:human_gvhmr_gmr}
\end{figure}

\subsection{Human Motion Representation}

Fig.~\ref{fig:human_gvhmr_gmr} shows the motion representation used in our implementation. Inputs may come from monocular reconstruction, RGB-D pose estimation, or an existing 3D motion file. At minimum, we require the root trajectory, torso orientation, and wrist trajectories. Object position and hand-object contact timing can further inform allocation in manipulation demonstrations. We normalize the input to a robot-centric frame by removing global camera drift, converting coordinate conventions, and estimating the initial facing direction. Our experiments use precomputed SMPL-style motion~\cite{loper2015smpl,shen2024gvhmr}, but the allocator only requires the normalized representation and is not tied to a specific estimator.

\subsection{Task-Conditioned Motion Allocator}

The allocator combines human motion cues with a lightweight task label, such as walking, reaching, bending, or picking. It is an interpretable optimization and repair module, not a learned semantic model. The task label resolves cases in which similar body motions imply different robot actions. Phase gating further prevents transient or irrelevant cues from activating the wrong subsystem. The resulting base, lift, and arm responsibilities form a robot-independent intermediate sequence.

\textbf{Locomotion cue:} Horizontal root displacement and yaw change drive the base. Under a locomotion condition, walking-like root motion receives a high base prior, while human arm swing is down-weighted as a manipulation target.

\textbf{Reach-height cue:} Hand height, torso pitch, and shoulder-to-hand displacement determine whether to move the vertical lift. A robot without a human-like waist cannot directly reproduce torso lowering. The allocator instead lowers the lift when available and replans the arm pose. It also stabilizes monocular root height so that camera or pose-estimation drift does not trigger unintended lift motion.

\textbf{Manipulation cue:} Wrist trajectories and hand orientations drive the arms. For manipulation tasks, the allocator increases hand-tracking gains and uses the base or lift when needed to keep the target reachable.

The intermediate motion file stores phase labels, base/lift/arm allocation weights, the base target trajectory, the lift reference, and hand targets. We treat the allocated planar trajectory as geometry that must be preserved. Heading stabilization and temporal repair may change its timing, but not its net displacement or cumulative progress. This representation separates motion interpretation from robot-specific realization.

\subsection{GMR-Based Motion Realization}

Given allocation weights $z_t$, the realization module solves for robot states and command trajectories. It first converts the intermediate states to GMR-compatible frames and applies GMR to the upper body. The allocated planar base pose and lift reference are then inserted into the mobile-base humanoid trajectory. We also propagate lift displacement to the arm-base target to maintain consistent end-effector motion.

We formulate realization as a constrained task-space problem. At each frame, the relevant subsystems receive task-space targets. A differential IK or quadratic programming (QP) solver then computes velocities that reduce task error while satisfying the constraints:
\begin{align}
    \min_{\dot{q}_t} \quad &
    \sum_i w_i \|J_i(q_t)\dot{q}_t - \dot{x}^{*}_{i,t}\|_2^2
    + \lambda \|\dot{q}_t\|_2^2 \\
    \text{s.t.} \quad &
    q_{\min} \le q_t + \Delta t \dot{q}_t \le q_{\max}, \\
    & \dot{q}_{\min} \le \dot{q}_t \le \dot{q}_{\max}.
\end{align}
Here, $J_i$ denotes the Jacobian for task $i$, such as hand position or orientation, and the allocation variables modulate the weights $w_i$. The planar base and lift are not treated as human-like links for one-to-one matching. GMR handles model-based upper-body retargeting; the allocator and adapter account for mobile-base components absent from standard humanoid retargeters.

\subsection{Base-Preserving Temporal Repair}

Framewise realization alone does not enforce consistency between consecutive robot states. Abrupt changes may arise when a target falls outside the workspace or the IK solution switches branches. We apply subsystem-specific temporal repair after GMR to smooth these changes without discarding the allocated base path. A minimum-jerk blend reconstructs base heading and suppresses unintended translation during dominant turns. We redistribute the removed translation over valid non-turn intervals, preserving net displacement and cumulative path progress. Motion-timed soft decoupling follows the turn timing of the human-derived motion rather than imposing a fixed ``turn first'' schedule. Speed retiming then reduces the effective frame rate by
\begin{equation}
    s=\max\!\left(1,\frac{\max_t\|v^{\mathrm{ref}}_t\|}{v_{\max}},
    \frac{\max_t|\omega^{\mathrm{ref}}_t|}{\omega_{\max}}\right),
    \qquad f_{\mathrm{out}}=\frac{f_{\mathrm{in}}}{s},
\end{equation}
which changes execution duration while retaining the repaired spatial path.

For the lift, we separate intentional bending and reaching cues from pose-estimation noise. Gating and velocity limits prevent monocular root-height drift from producing spurious lift motion. For the arms, stable-start minimum-jerk trajectories replace noisy joint-space transitions. Low-object pick-like motions use a two-stage open-to-close trajectory coordinated with lift lowering.

\begin{figure*}[!t]
    \centering
    \includegraphics[width=\textwidth]{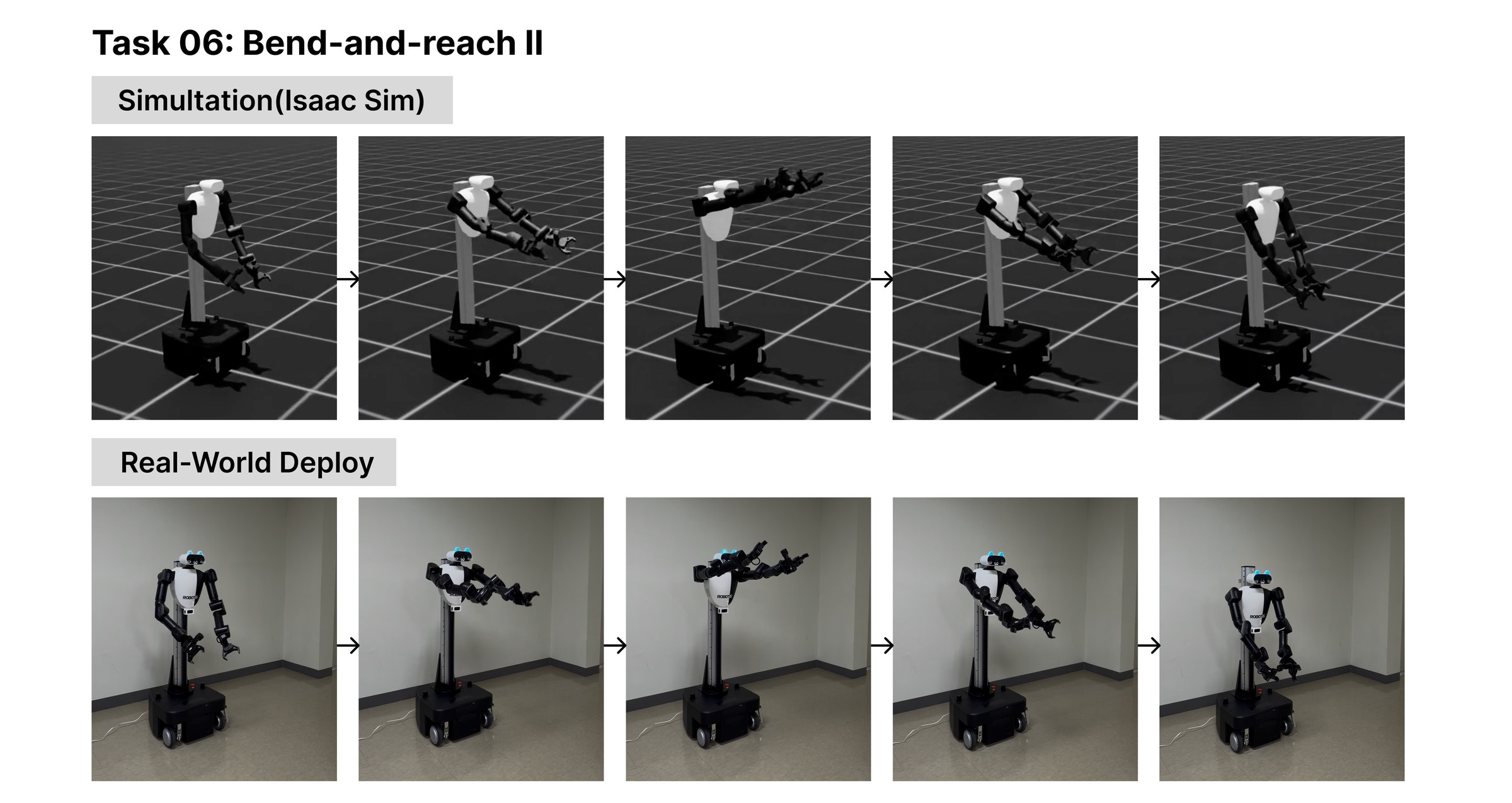}
    \caption{The same bend-and-reach reference executed in simulation and on the physical robot for Task 06. The sequence retains the intended lift and bilateral arm coordination in both environments.}
    \label{fig:bend-reach-execution}
\end{figure*}

\subsection{Command and Deployment Adapter}

We resample the repaired position reference to 50 Hz for simulation and command export. The adapter differentiates the planar base pose into $(v_x,v_y,\omega)$. When net displacement and yaw are negligible but the path-to-net ratio is high, it treats the trajectory as pose-estimation jitter and holds the base stationary. Otherwise, it applies moving-average smoothing, linear/angular deadbands, and separate acceleration slew limits. The hold, deadband, and slew checks run again immediately before command publication.

Playback speed and command magnitude are coupled to preserve path geometry. For playback factor $\alpha \in (0,1]$,
\begin{equation}
    t'_k=t_k/\alpha,\quad
    [v'_x,v'_y]=\alpha g_{\mathrm{lin}}[v_x,v_y],\quad
    \omega'=\alpha g_{\mathrm{ang}}\omega .
\end{equation}
This coupling prevents slower playback from increasing total displacement or yaw. Independent gains $g_{\mathrm{lin}}$ and $g_{\mathrm{ang}}$ support platform-specific calibration without tying lateral translation to rotation. The adapter remains separate from human-motion allocation and serves as the embodiment-specific interface to the robot command space.

\section{EXPERIMENTS}

Our experiments compare embodiment-aware allocation with subsystem-limited retargeting baselines. We evaluate the generated references through simulation replay and component ablations. These experiments assess whether the references can be loaded, replayed, and compared across decomposition baselines; they do not measure learned-controller performance.

\subsection{Target Platform}

The physical demonstration uses a ROBOTIS AI Worker (FFW-SG2). This mobile-base humanoid has an omnidirectional wheeled base, two 7-DoF arms, grippers, head joints, and a vertical lift. Its reference contains 31 joints: wheel drive and steering joints, both arms, grippers, head joints, and the lift joint. Realization and command export use the measured robot model, joint and velocity limits, lift range, and controller interface.

\subsection{Motion Categories}

We selected nine representative scenarios from a larger collection:
\begin{itemize}
    \item \textbf{Locomotion:} slow walk, run, side walk, and sideways-turn walk, expressed mainly through the base while suppressing task-irrelevant arm swing.
    \item \textbf{Bend-and-reach:} forward bending at different reach amplitudes, requiring coordinated lift and arm motion with little base drift.
    \item \textbf{Arm-and-base coordination:} a backward step with arms forward and a wide-arm lateral motion, combining base relocation with smooth arm timing.
    \item \textbf{Low pick-like motion:} a low-object pick pose, requiring lift lowering and a coordinated arm approach.
\end{itemize}

\subsection{Baselines}

We compare against the following baselines:
\begin{itemize}
    \item \textbf{Arm-only IK:} map human wrists to robot end-effectors without moving the base or lift.
    \item \textbf{Base-only root tracking:} map human root motion to base motion and keep arms near a nominal pose.
    \item \textbf{No-lift allocation:} preserve the generated base and arm trajectories while freezing the vertical lift. This isolates the role of lift assignment in bending and low-object motions.
    \item \textbf{Ours:} the proposed base-preserving allocation, GMR-based realization, and temporal repair stack.
\end{itemize}

\subsection{Metrics}

Our metrics separate \emph{semantic imitation} from \emph{reference executability}. Semantic imitation uses human-derived task targets extracted from GVHMR-reconstructed SMPL motion rather than our generated robot trajectory. We compare only task-relevant components: root displacement and speed for locomotion, wrist displacement and height for reaching, and torso-lowering cues converted to desired lift-height compensation. Since monocular absolute scale is imperfect, we normalize component errors within each scenario before averaging. The resulting costs support relative comparisons: lower values indicate that the robot reference retains more of the intended human cue.

We measure executability after converting each trajectory to a 50 Hz simulation reference. A replay is complete when it loads and runs without early termination. Body mean error is the average Cartesian discrepancy between the main robot body and its replay target. Anchor mean error measures consistency at the base and end-effector anchor frames. Tables~\ref{tab:human_continuous} and~\ref{tab:component_ablation} evaluate the allocation, while Table~\ref{tab:direct_playback} tests whether the same references are ready for simulation.

The reported quantities are:
\begin{itemize}
    \item hand trajectory error in the robot/world frame;
    \item approach-direction error between human and robot hand motion;
    \item base trajectory error for locomotion-dominant segments;
    \item lift-height proxy error for vertical reach segments;
    \item joint-limit, velocity-limit, and workspace violation rates;
    \item command smoothness measured by velocity and acceleration variation;
    \item policy-free direct-playback completion in simulation, including whether a sequence terminates early.
\end{itemize}

\section{REFERENCE VALIDATION}

Our implementation reads reconstructed 3D human motion and produces task-conditioned base, lift, and arm allocations. It realizes the upper body through GMR, repairs the trajectory while preserving the base path, and renders kinematic previews. The 50 Hz command adapter exports the same references for execution. In Fig.~\ref{fig:bend-reach-execution}, Task 06 uses one reference in both simulation and the physical robot; the intended lift and bilateral arm coordination is retained in both. The remaining results cover the larger simulation set.

The evaluation set comprises nine generated trajectories spanning locomotion, bending and reaching, base-arm coordination, and low-pick-like motion. We examine continuous error against human-derived task targets, component-wise ablations, and policy-free simulation playback.

Table~\ref{tab:human_continuous} reports normalized task-space costs against the human-derived target. These costs are not binary success labels, and the generated motion is not used as ground truth. Methods that retain base motion are expected to perform similarly on locomotion. Manipulation and reaching, by contrast, reveal whether the base, lift, and arms are coordinated. We report grouped means for readability and retain the full scenario-wise matrix with the experiment artifacts. The full method matches the base-preserving baselines on locomotion and achieves the lowest grouped mean across all nine scenarios.

\begin{table}[t]
\caption{Human-derived continuous task-space evaluation on scenarios 01--09. Targets are extracted from GVHMR-reconstructed SMPL motion rather than from our generated reference. Entries are normalized costs; lower is better.}
\label{tab:human_continuous}
\centering
\scriptsize
\begin{tabular}{lccc}
\toprule
Method & Locomotion & Manip./reach & All \\
 & 01--04 & 05--09 & 01--09 \\
\midrule
Ours & \textbf{0.355} & \textbf{0.295} & \textbf{0.322} \\
Arm-only & 1.000 & 0.566 & 0.759 \\
Base-only & \textbf{0.355} & 0.868 & 0.640 \\
No-lift & \textbf{0.355} & 0.440 & 0.402 \\
\bottomrule
\end{tabular}
\end{table}

Table~\ref{tab:component_ablation} breaks the same evaluation down by component. Arm-only loses locomotion and lift intent. Base-only preserves root motion but not wrist behavior, while no-lift retains much of the arm and base motion but misses bend-to-lift compensation. No restricted baseline covers every task group. The full method matches their locomotion performance while retaining both wrist behavior and lift compensation, supporting coordinated allocation across the base, lift, and arms.

\begin{table}[t]
\caption{Component-wise human-target ablation. Locomotion root is averaged over scenarios 01--04, wrist metrics over scenarios 05--09, and lift shortfall over scenarios 05, 06, and 09. Lower is better.}
\label{tab:component_ablation}
\centering
\scriptsize
\resizebox{\linewidth}{!}{%
\begin{tabular}{lccccl}
\toprule
Method & Loc. root & Wrist & Wrist-$z$ & Lift short. & Primary gap \\
 & (m) & (m) & (m) & (m) & \\
\midrule
Ours & \textbf{0.641} & \textbf{0.346} & \textbf{0.143} & \textbf{0.012} & balanced \\
Arm-only & 1.610 & 0.352 & 0.164 & 0.125 & base/lift \\
Base-only & \textbf{0.641} & 1.726 & 0.643 & 0.125 & wrist/lift \\
No-lift & \textbf{0.641} & \textbf{0.346} & 0.164 & 0.125 & lift \\
\bottomrule
\end{tabular}
}
\end{table}

The repair stages address distinct reference-generation failures. Base repair preserves human-derived planar progress while correcting lateral-axis interpretation, heading, and speed. Its motion-timed turn/translation gate avoids simultaneous high linear and angular commands without inserting a long pause before turning. Lift gating rejects monocular height drift while retaining intentional lowering. In the inspected wide-arm sample, replacing per-frame GMR arm blending with a minimum-jerk trajectory reduced peak joint speed from 4.789 rad/s to 2.201 rad/s and peak acceleration from 28.128 rad/s$^2$ to 3.755 rad/s$^2$. For low-pick motion, the two-stage open-close trajectory reduced hand separation from approximately 0.89 m to 0.52 m as the lift lowered to -0.18 m.

The command adapter handles a failure that is not visible in kinematic video. If playback is slowed by $1/\alpha$ while base velocity remains unchanged, integrated translation and yaw increase by the same factor. We therefore apply $\alpha$ to both linear and angular velocity and retain separate calibration gains. The adapter also freezes jitter-like, near-stationary base paths and limits command acceleration. These operations leave the task-space scores in Tables~\ref{tab:human_continuous} and~\ref{tab:component_ablation} unchanged, but preserve the reference geometry at the simulator/robot interface.

Table~\ref{tab:direct_playback} reports policy-free replay of the generated references. We exported all nine trajectories at 50 Hz; each replay completed without early termination, for 2947 total steps. We omit raw joint-position error because wheel drive joints dominate this aggregate. Body and anchor consistency provide more informative diagnostics at this stage.

\begin{table}[t]
\caption{Policy-free simulation replay validation. Completion means no early termination during reference playback, not learned-policy success. Lower errors are better.}
\label{tab:direct_playback}
\centering
\scriptsize
\resizebox{\linewidth}{!}{%
\begin{tabular}{lccccc}
\toprule
Group & Sc. & Completed & Body mean (m) & Anchor mean (m) & Steps \\
\midrule
Locomotion & 4 & 4/4 & 0.012 & 0.175 & 1427 \\
Bend/reach & 2 & 2/2 & 0.024 & 0.020 & 551 \\
Base+arms & 2 & 2/2 & 0.028 & 0.048 & 654 \\
Low pick & 1 & 1/1 & 0.029 & 0.032 & 315 \\
\midrule
All 01--09 & 9 & 9/9 & 0.020 & 0.097 & 2947 \\
\bottomrule
\end{tabular}%
}
\end{table}

\section{DISCUSSION AND LIMITATIONS}

Our formulation is less anthropomorphic than standard humanoid retargeting, which suits mobile-base robots but introduces ambiguity. The same human motion may correspond to several robot command sequences. A far reach, for example, can be expressed by extending the arm, driving the base, raising the lift, or combining all three. The allocator must choose among these alternatives while respecting both feasibility and natural robot behavior.

The study has four main limitations. First, the experiments use precomputed 3D human motion; comparing alternative reconstruction methods remains future work. Second, the allocator is interpretable and partly scenario-conditioned. Broader baselines, including direct GMR without allocation, and held-out evaluation would better characterize generalization without per-sequence tuning. Third, direct playback verifies reference consistency and simulator compatibility but not disturbance rejection. Finally, the physical-robot result is qualitative, and broader hardware evaluation should measure completion and tracking error systematically.

Learning-based tracking is a downstream direction rather than a claim of this work. A supervised tracker or PPO policy could assess how well the references withstand dynamics and perturbations. Our results instead support explicit allocation before retargeting: GMR-based realization yields replayable mobile-base references and exposes the limitations of subsystem-restricted mappings.

\section{CONCLUSION}

We introduced task-conditioned embodiment-aware motion allocation for human video imitation on mobile-base humanoids. Rather than matching joints directly, the method assigns motion to the base, lift, and arms, then repairs heading, timing, and subsystem trajectories while preserving base-path geometry. GMR realizes the upper body, and a 50 Hz adapter handles stationary-base suppression, slew limits, and time-consistent linear/angular scaling. Across nine scenarios, our references were compared with subsystem-restricted baselines and replayed without early termination. We also executed one reference on the physical robot as a qualitative demonstration. Future evaluation will add broader retargeting baselines, held-out scenarios, and quantitative physical-robot measurements.

\bibliographystyle{IEEEtran}
\bibliography{references}

\end{document}